%% file: main.tex
\PassOptionsToPackage{table,dvipsnames}{xcolor}
\documentclass[10pt,twocolumn,letterpaper]{article}

\usepackage[pagenumbers]{iccv}
\input{preamble}

\definecolor{iccvblue}{rgb}{0.21,0.49,0.74}
\usepackage[pagebackref,breaklinks,colorlinks,allcolors=iccvblue]{hyperref}

\title{What's Making That Sound Right Now? Video-centric Audio-Visual Localization}

\author{
Hahyeon Choi \quad Junhoo Lee \quad Nojun Kwak \\
Seoul National University \\
{\tt\small \{hahyeon.choi, mrjunoo, nojunk\}@snu.ac.kr}
}

\begin{document}
\maketitle
\input{sec/0_abstract}   
\input{sec/1_introduction}
\input{sec/2_related_work}
\input{sec/3_benchmark}
\input{sec/4_approach}
\input{sec/5_experiments}
\input{sec/6_conclusion}
\input{sec/acknowledgements}

{
    \small
    \bibliographystyle{ieeenat_fullname}
    \bibliography{main}
}

\end{document}

%% file: preamble.tex
\usepackage{xcolor}

\definecolor{check}{rgb}{0.333, 0.755, 0.545}
\definecolor{tcheck}{rgb}{1.0, 0.65, 0.0} 
\newcommand{\cmark}{\ding{51}}
\newcommand{\xmark}{\ding{55}}

\usepackage{kotex}

\usepackage{pifont}
\usepackage{makecell}

\usepackage{rotating}

\usepackage{array}
\newcolumntype{L}[1]{>{\raggedright\let\newline\\\arraybackslash\hspace{0pt}}m{#1}}
\newcolumntype{C}[1]{>{\centering\let\newline\\\arraybackslash\hspace{0pt}}m{#1}}
\newcolumntype{R}[1]{>{\raggedleft\let\newline\\\arraybackslash\hspace{0pt}}m{#1}}

\usepackage{algorithm}
\usepackage{algpseudocode}

\usepackage{threeparttable}

%% file: sec/0_abstract.tex
\begin{abstract}

Audio-Visual Localization (AVL) aims to identify sound-emitting sources within a visual scene. However, existing studies focus on image-level audio-visual associations, failing to capture temporal dynamics. Moreover, they assume simplified scenarios where sound sources are always visible and involve only a single object. To address these limitations, we propose AVATAR, a video-centric AVL benchmark that incorporates high-resolution temporal information. AVATAR introduces four distinct scenarios -- Single-sound, Mixed-sound, Multi-entity, and Off-screen -- enabling a more comprehensive evaluation of AVL models. Additionally, we present TAVLO, a novel video-centric AVL model that explicitly integrates temporal information. Experimental results show that conventional methods struggle to track temporal variations due to their reliance on global audio features and frame-level mappings. In contrast, TAVLO achieves robust and precise audio-visual alignment by leveraging high-resolution temporal modeling. Our work empirically demonstrates the importance of temporal dynamics in AVL and establishes a new standard for video-centric audio-visual localization.\footnote{\url{https://hahyeon610.github.io/Video-centric_Audio_Visual_Localization/}}

\end{abstract}

%% file: sec/1_introduction.tex
\vspace{-0.5cm}

\section{Introduction}
\label{sec:intro}

Humans and animals perceive the world by integrating multiple sensory modalities. In particular, vision and hearing complement each other, enabling the localization and identification of sound sources across a broad spatial field. This capability is essential for filtering relevant auditory information, such as distinguishing a speaker in a crowded environment or detecting a predator’s location from its roar. Inferring spatial information from multimodal cues has broad applications in areas such as robot navigation, AR/VR, and video analysis.

Audio-Visual Localization (AVL) seeks to replicate this perceptual ability by identifying sound sources within a visual scene. A core approach in AVL research leverages the natural co-occurrence of audio and visual signals, using self-supervised learning to align these modalities and extract joint embeddings. This paradigm has led to significant advancements in AVL~\cite{morgado2020learning, tian2018audio, zhao2018sound, chen2021localizing, afouras2020self, 7952687}, enabling models to effectively utilize large-scale web video datasets without requiring manual annotations. However, despite this progress, existing AVL research faces two key limitations, largely stemming from the structural constraints of current AVL benchmarks.

The first limitation is that AVL research primarily focuses on image-level audio-visual associations rather than extending to video-based analysis. Most benchmarks~\cite{Senocak_2018_CVPR, Chen21} adopt an annotation approach where annotators watch the entire video and label all sound-emitting objects in a single frame, effectively treating that frame as representative of the entire video. Consequently, existing methods~\cite{senocak2023sound, hu2019deep, sun2023learning, park2024can} operate on single-frame inputs, neglecting temporal dynamics. While this approach assesses spatial understanding within static images, it fails to capture temporal variations. Real-world AVL tasks require tracking moving sound sources and handling dynamic changes over time, making spatiotemporal modeling essential for robust performance.

% ------
\input{table/comparision_with_avl_benchmarks}
% ------

The second limitation is the oversimplified assumptions in existing AVL benchmarks~\cite{zhou2022avs, Senocak2024AligningSA, Senocak_2018_CVPR, Chen21}. These benchmarks assume that sound-emitting objects are always visible and typically involve only a single active source. However, real-world scenarios often feature multiple simultaneous sound sources, and sound-emitting objects may be outside the visual field. These constraints limit model generalization to complex audio-visual environments. Although some studies attempt to address these issues, they primarily rely on negative pair construction using mismatched image-audio pairs~\cite{mo2022SLAVC} or synthetic mixed-audio generation~\cite{Senocak2024AligningSA}, providing only partial solutions.

To overcome these limitations, we propose advanced \textbf{A}udio-\textbf{V}isual localiz\textbf{A}tion benchmark for a spatio-\textbf{T}empor\textbf{A}l pe\textbf{R}spective in video (AVATAR), a video-centric AVL benchmark that evaluates models on their ability to localize sound sources with high temporal resolution. As summarized in \cref{tab:benchmarks}, AVATAR introduces four key evaluation scenarios that reflect real-world complexity: Single-sound, Mixed-sound, Multi-entity, and Off-screen.

Furthermore, we introduce \textbf{T}emporal-aware \textbf{A}udio-\textbf{V}isual \textbf{L}ocalization model for fine-grained vide\textbf{O} understanding (TAVLO), a novel AVL model that explicitly incorporates temporal information. Experimental results demonstrate that existing methods struggle with tracking temporal changes, as they rely on global audio features and static frame mappings. In contrast, TAVLO is robust to temporal variations, leveraging high-resolution temporal information for precise audio-visual alignment. Our model effectively localizes sound sources even in complex environments, demonstrating significant improvements over prior approaches.

%% file: table/comparision_with_avl_benchmarks.tex
\begin{table*}[!t]
\begin{center}
\caption{
Comparison of Audio-Visual Localization Datasets. A comparison of existing AVL datasets in terms of dataset scale, annotation type, and the scenarios they cover, including Single-sound, Mixed-sound, Multi-entity, and Off-screen. $\S$ Statistics for AVSBench and AVSBench-Semantic are based on the test set due to supervised use.
}
\vspace{-3mm}
\resizebox{\textwidth}{!}{
\renewcommand{\arraystretch}{1.35}
\begin{tabular}{lcccccccccc}
\Xhline{3\arrayrulewidth}
 &  &  &  &  &  & & \multicolumn{4}{c}{Scenario} \\
\cmidrule(r){8-11}
\textbf{Dataset} & \# Videos & \# Frames & \# Categories & Avg. Length & Annotation & Annotation type & Single-sound & Mixed-sound & Multi-entity & Off-Screen \\
\hline \hline
Flickr-SoundNet~\cite{Senocak_2018_CVPR} & 5,000 & 5,000 & 50 & 20.0s & bbox & Image & {\color{check}\cmark} & \xmark & \xmark & \xmark \\
VGGSS~\cite{Chen21}& 5,158 & 5,158 & 221 & 10.0s & bbox & Image &  {\color{check}\cmark} & \xmark & \xmark & \xmark \\
Epic Sound Object~\cite{huang2023egocentric} & 3,172 & 9,196 & 30 & 1.0s & bbox & Image & {\color{check}\cmark} & \xmark & \xmark & \xmark \\
Extended Flickr-SoundNet~\cite{mo2022SLAVC} & 292 & 292 & 50 & 15.4s & bbox & Image & \xmark & \xmark & \xmark & {\color{check}\cmark} \\
Extended VGGSS~\cite{mo2022SLAVC} & 5,537 & 5,537 & 221 & 10.0s & bbox & Image & \xmark & \xmark & \xmark & {\color{check}\cmark} \\
IS3~\cite{Senocak2024AligningSA}  & 3,240 & 3,240 & 118 & 10.0s & \makecell[c]{bbox \& \\ instance segm.} & Image & \xmark & {\color{check}\cmark} & \xmark & \xmark \\
$^\S$AVSBench~\cite{zhou2022avs} & 804 & 4,020 & 23 & 5.0s & segm. & Chunk & {\color{check}\cmark} & {\color{check}\cmark} & \xmark & \xmark \\
$^\S$AVSBench-Semantic~\cite{zhou2023avss} & 1,554 & 11,520 & 70 & 7.4s & semantic segm. & Chunk & {\color{check}\cmark} & {\color{check}\cmark} & \xmark & \xmark \\
$\mathbf{AVATAR}$ (Ours) & 5,000 & 24,266 & 80 & 10.0s & \makecell[c]{bbox \& \\ instance segm.} & Video & {\color{check}\cmark} & {\color{check}\cmark} & {\color{check}\cmark} & {\color{check}\cmark} \\
\Xhline{3\arrayrulewidth}
\end{tabular}
}
\label{tab:benchmarks}
\end{center}
\vspace{-0.6cm}
\end{table*}

%% file: sec/2_related_work.tex
\section{Related Work}
\label{sec:relwork}

\subsection{Audio-Visual Localization}

AVL has primarily been studied through self-supervised learning, leveraging the natural co-occurrence of auditory and visual signals. Various approaches have been proposed to enhance AVL performance~\cite{senocak2023sound, sun2023learning, hu2019deep}. EZ-VSL~\cite{mo2022localizing} formulates AVL as a multiple instance learning~\cite{Maron1997AFF} problem, introducing a cross-modal multiple-instance learning loss to improve training. SSL-TIE~\cite{liu2022exploiting} enhances self-supervised localization by incorporating transformation invariance and equivariance. ACL-SSL~\cite{park2024can} integrates CLIP~\cite{radford2021learning} to explore multimodal learning in AVL. To address the limitations of purely self-supervised methods, DMT~\cite{guo2023dual} employs semi-supervised learning to improve generalization, while SLAVC~\cite{mo2022SLAVC} explores weakly supervised strategies for precise localization with limited annotations. Beyond single-source scenarios, several studies have extended AVL to multi-source environments~\cite{qian2020multiple, kim2024learning, hu2022mix, mo2023audio}, broadening its applicability to more complex auditory scenes.

% ------
\begin{figure*}[t]
    \centering
    \includegraphics[width=\textwidth]{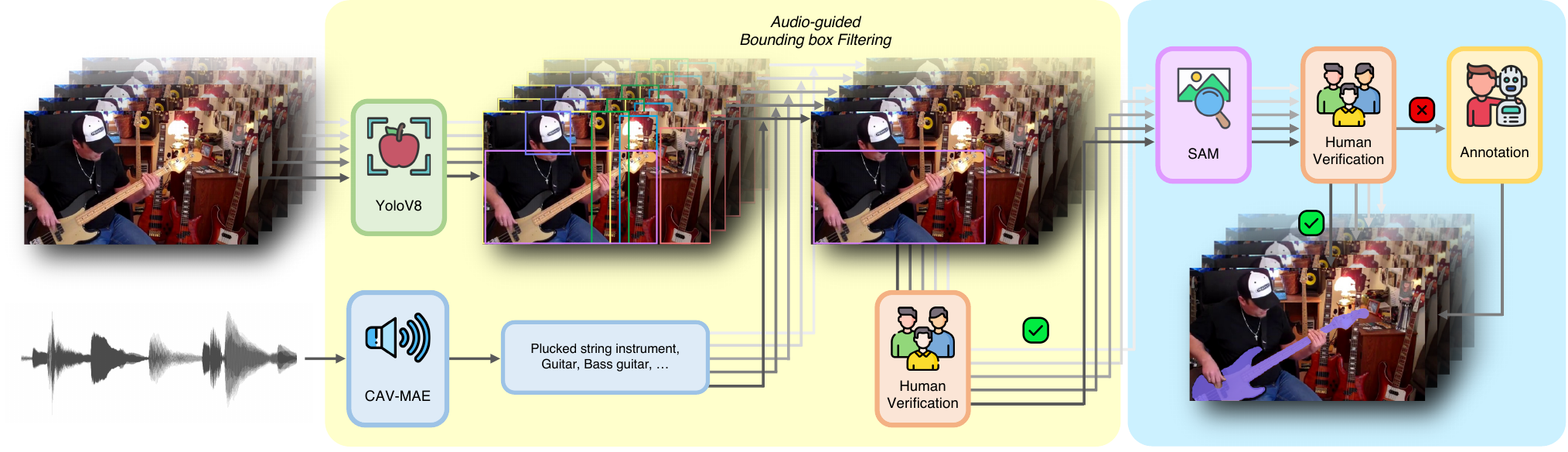}
    \caption{Model-Driven Labeling and Verification Process. CAV-MAE \cite{gong2023contrastive} and YoloV8 \cite{reis2023real} generate candidate bounding boxes with audio class predictions, refined through Audio-guided Bounding box Filtering. Human verification filters the regions before SAM \cite{kirillov2023segment} performs instance segmentation, followed by a final verification for accuracy.}
    \label{fig:pipeline}
    \vspace{-0.5cm}
\end{figure*}
% ------

\subsection{Audio-Visual Localization Benchmarks}

The most widely used datasets, Flickr-SoundNet~\cite{Senocak_2018_CVPR} and VGGSS~\cite{Chen21}, are derived from SoundNet~\cite{aytar2016soundnet} and VGGSound~\cite{chen2020vggsound}, respectively. In the egocentric domain, Epic Sound Object~\cite{huang2023egocentric} introduces an AVL benchmark that considers viewpoint variations and occlusion challenges arising from wearer-environment interactions. Other benchmarks explore misalignment scenarios and multi-source settings. Extended datasets~\cite{mo2022SLAVC} are designed to evaluate model performance under audio-visual misalignment by generating synthetic negative samples through cross-pairing audio and video frames from different sources. IS3~\cite{Senocak2024AligningSA} focuses on mixed-sound environments, combining VGGSound audio and generating synthetic images using Stable Diffusion. However, these benchmarks primarily provide image-level labels for a single frame (or up to three frames in Epic Sound Object), limiting their ability to assess temporal dynamics. To address this limitation, AVSBench~\cite{zhou2022avs} and AVSBench-Semantic~\cite{zhou2023avss} introduce the Audio-Visual Segmentation task, highlighting the inability of conventional AVL benchmarks to capture the shape and extent of sound-emitting objects. AVSBench annotates one-second clips with segmentation masks for visible sounding objects. However, this clip-based annotation approach remains restricted, as each clip is still represented by a single frame. \cref{tab:benchmarks} summarizes major AVL benchmarks.

%% file: sec/3_benchmark.tex
\section{Benchmark: AVATAR}
\label{sec:benchmark}

AVATAR addresses the limitations of existing benchmarks by adopting a video-centric approach, providing a more precise evaluation framework that reflects real-world complexity. To achieve this, Section~\ref{subsec:annotation_pipeline} introduces a semi-automatic annotation pipeline, which maintains high temporal resolution while reducing labeling costs and ensuring annotation quality. Subsequently, Section~\ref{subsec:scenario_definitions} defines the four key evaluation scenarios considered in AVATAR and discusses the essential factors for model assessment in each setting.

\subsection{Semi-Automatic Annotation Pipeline}
\label{subsec:annotation_pipeline}

Annotating sound-emitting instances with gold-standard labels is labor-intensive and costly due to the need for human intervention. To address this, the semi-automatic annotation pipeline integrates deep learning models to minimize manual effort while maintaining high annotation quality. The pipeline consists of three stages. First, \cref{subsubsec:candidate_videos} describes the selection process for candidate raw videos. Next, \cref{subsubsec:automatic_sampling} introduces an automated sampling strategy to identify key video clips and frames that require annotation, ensuring efficiency while mitigating sampling bias. This step involves two phases: (i) Clip Sampling, which extracts meaningful video segments, and (ii) Frame Sampling, which selects key frames for annotation. Finally, \cref{subsubsec:model-driven_labeling} details a model-assisted labeling process, where off-the-shelf models generate initial annotations, allowing human annotators to focus on verification and refinement. \cref{fig:pipeline} provides an overview of the process.

\subsubsection{Obtaining Candidate Videos}
\label{subsubsec:candidate_videos}

The dataset is constructed using VGGSound~\cite{chen2020vggsound}, a large-scale video dataset sourced from YouTube under the Creative Commons Attribution 4.0 International License. Using YouTube API, we retrieve the original videos based on their YouTube IDs, collecting approximately 70k raw videos. To ensure quality and consistency, we apply a filtering process with constraints on resolution (640$\times$360), frame rate (20--40 fps), duration ($\ge$10s), and bitrate ($\ge$100 bps), resulting in 39k candidate videos for annotation.

% ------
\begin{figure*}[t]
    \centering
    \includegraphics[width=\textwidth]{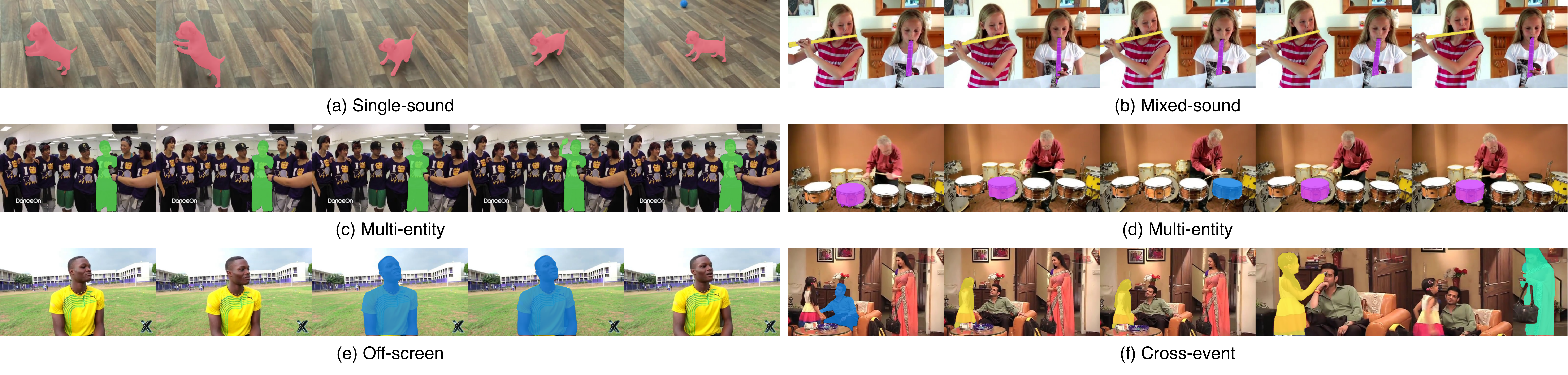}
    \vspace{-0.7cm}
    \caption{Four Scenarios and Cross-event Subset Examples. (a) Single-sound: A single source (e.g., a barking dog). (b) Mixed-sound: Overlapping sources (e.g., flute and clarinet). (c, d) Multi-entity: Multiple distinct sources, such as an interview in a crowd (c) or multiple drums (d). (e) Off-screen: The source is outside the frame (e.g., a speaker behind the camera). (f) Cross-event: The active sound source changes over time.
}
    \label{fig:senarios}
    \vspace{-0.5cm}
\end{figure*}
% ------

\subsubsection{Automatic Clip and Frame Sampling}
\label{subsubsec:automatic_sampling}

\paragraph{Clip Sampling} This stage extracts video segments containing meaningful audio. The Root Mean Square (RMS) energy of the audio signal is computed to distinguish between silence and audio-active regions, with a threshold of 0.01 applied at 1s intervals. Using a sliding window approach, we identify 10s segments where audio activity occurs at least five times, selecting them as candidate clips. The final annotation clips are then randomly sampled from these candidates.

\vspace{-2mm}
\paragraph{Frame Sampling} Once a clip is selected, this step determines the frames to be labeled. We define audio events based on RMS energy peaks within a $\pm$0.1s window. Given the filtering constraints in \cref{subsubsec:candidate_videos}, each 0.2s audio event contains at least four frames. To minimize motion blur, we select the sharpest frame by applying a Laplacian filter, a high-pass filter that enhances edge contrast. Additionally, up to five frames per clip are sampled based on audio event intensity, ensuring a balanced temporal distribution by excluding frames from overlapping events within $\pm$0.7s.

\subsubsection{Model-Driven Labeling and Verification}
\label{subsubsec:model-driven_labeling}

\paragraph{Target Category Selection} To effectively utilize off-the-shelf models, we first determine the target categories for annotation. We analyze existing audio-visual datasets, including VGGSound \cite{chen2020vggsound} (300 classes), OpenImageV7 \cite{OpenImages2} (600 classes), and AudioSet~\cite{7952261} (527 classes). Categories where the sound source is ambiguous (e.g., people crowd, wind noise) are excluded. Ultimately, we select 80 target categories covering a broad range of real-world domains.

\vspace{-3mm}
\paragraph{Audio-Guided Bounding-box Annotation} We employ YoloV8 \cite{reis2023real}, trained on OpenImageV7 \cite{OpenImages2}, for object detection, and CAV-MAE \cite{gong2023contrastive}, trained on AudioSet \cite{7952261}, for audio classification. First, among the bounding boxes created by YoloV8, instances that did not correspond to the target categories are filtered out. Next, CAV-MAE’s classification results guide the selection of bounding boxes associated with active sound sources, implementing an Audio-Guided Bounding Box Filtering strategy. Human annotators verify and filter these bounding boxes by reviewing the corresponding video segment ($\pm\textbf{0.05}$s around the frame) to ensure that the labeled instance is indeed producing sound at that moment.

\vspace{-2mm}
\paragraph{Model-Assisted Instance Segmentation} Verified bounding boxes are used as prompts for SAM~\cite{kirillov2023segment} to generate instance segmentation masks automatically. Human verification ensures segmentation accuracy, and only in cases where the mask was not well generated, it was supplemented with human annotation under the guidance of SAM.

\subsection{Scenario Definitions}
\label{subsec:scenario_definitions}

A key distinction of our benchmark is the introduction of instance-level scenario definitions, enabling fine-grained evaluation of AVL models under diverse real-world conditions. We define four distinct scenarios, each addressing a specific challenge in audio-visual localization. Visual illustrations are provided in \cref{fig:senarios}.

\vspace{-0.3cm}
\paragraph{Single-sound} represents the simplest case, where only one instance within the frame emits a clear and distinct audio signal. By eliminating complex auditory interference, it evaluates a model’s ability to learn and localize a one-to-one audio-visual correspondence, serving as a baseline before progressing to more complex settings.

\vspace{-0.3cm}
\paragraph{Mixed-sound} involves multiple concurrent audio sources, requiring the model to discriminate and associate sounds with their correct visual sources. It includes three variations: (1) multiple instances of the same category, (2) instances from different categories, and (3) partially off-screen sound sources. This setting tests the model’s ability to perform auditory scene separation and resolve multiple simultaneous audio-visual associations.

\vspace{-0.3cm}
\paragraph{Multi-entity} challenges the model to identify the actual sound-emitting instance among multiple visually similar objects, as first introduced in AVATAR. For example, in a scene with multiple people, only one may be speaking, requiring the model to pinpoint the correct speaker. Unlike conventional AVL approaches that rely solely on single-frame associations, this scenario necessitates spatiotemporal reasoning to distinguish active from passive entities within the same visual category.

\vspace{-0.3cm}
\paragraph{Off-screen} 
Unlike cameras with a limited field of view, microphones record omnidirectional audio, often including off-screen sound sources. This scenario evaluates a model’s ability to avoid false positives when no visible sound-emitting object is present, serving as a robustness check to ensure accurate localization only within the visual frame.

%% file: sec/4_approach.tex
\vspace{-1mm}
\section{Approach: TAVLO}
\label{sec:approach}

\begin{figure}
    \centering
    \includegraphics[width=\linewidth]{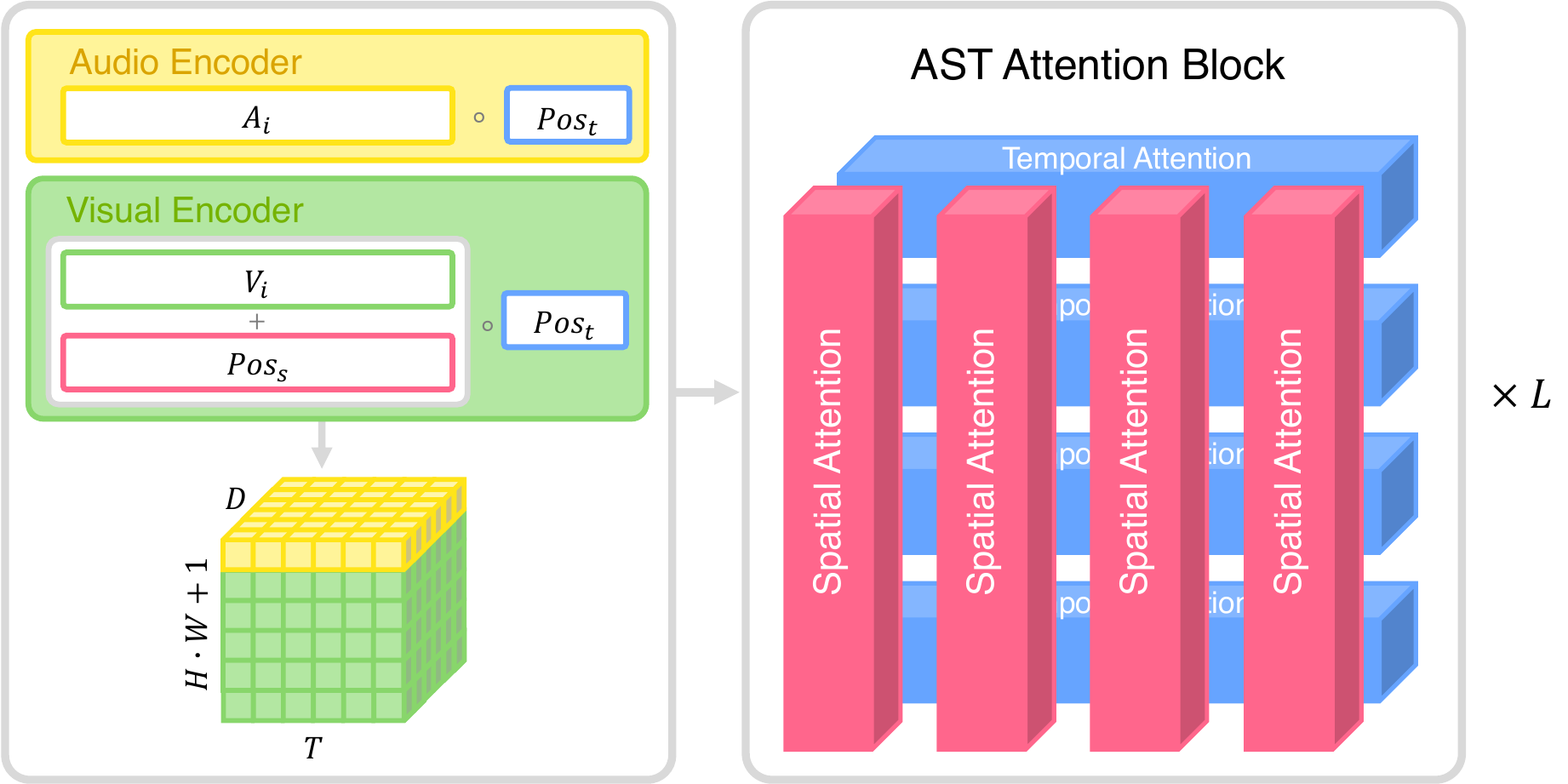}
    \caption{Overview of Our Approach. The audio and visual encoders extract features, applying positional encoding in a modality-specific manner. The AST Attention Block then applies stacked spatial and temporal attention layers, repeated $L$ times.
    }
    \label{fig:ast_block}
    \vspace{-0.6cm}
\end{figure}

Existing AVL studies primarily focus on audio-visual associations, overlooking temporal dynamics in video. To address this limitation, we propose a spatiotemporal AVL model that effectively integrates spatial and temporal information. The overall architecture is illustrated in \cref{fig:ast_block}.

Attention mechanisms have achieved state-of-the-art performance in both NLP and Vision by enabling adaptive feature selection and capturing long-range dependencies. Motivated by this, we employ attention-based modeling to effectively integrate audio-visual cues and localize the sound-emitting object at each time step. However, the quadratic complexity of self-attention poses a significant computational challenge when applied directly to flattened video features. To overcome this, we adopt a factorized attention strategy~\cite{Weissenborn2020Scaling, arnab2021vivit, bertasius2021space} and design the Audio-Spatial-Temporal (AST) Attention Block, enabling efficient processing of audio-visual sequences.

This section details our approach. \cref{approach:modality-specific_feature_encoding} introduces modality-specific feature encoding for time-aware audio-visual localization. \cref{approach:audio-spatial-temporal_attention_block} describes the AST Block architecture, and \cref{approach:training_objective} discusses the training objective.

\subsection{Modality-Specific Feature Encoding}
\label{approach:modality-specific_feature_encoding}

A given video $X$ consists of visual frames and audio, represented as $V \in \mathbb{R}^{T \times H_v \times W_v \times 3}$ and $A \in \mathbb{R}^{H_a \times W_a \times 1}$, respectively. Here, $V$ corresponds to a sequence of RGB frames, while $A$ represents the spectrogram of the raw audio waveform. $T$ denotes the total number of frames in the video. $H_v, W_v$ denote the height and width of video frames, while $H_a, W_a$ represent the frequency and time dimensions of the spectrogram.

Conventional AVL models typically adopt a two-stream neural network encoder, extracting global audio and single-frame visual features independently. However, this approach fails to capture temporal dynamics. To remedy this, we retain the two-stream encoder structure but introduce a time-specific encoding strategy that encodes modality-specific features at each time step.

\vspace{-4mm}
\paragraph{Feature Extraction} To encode visual and audio features, we employ a visual encoder $f_v$ and an audio encoder $f_a$. For visual feature extraction, we use ResNet-18~\cite{he2016deep} as $f_v$, following prior works~\cite{mo2022SLAVC, sun2023learning, mo2022localizing, senocak2023sound}. Given $T$ frames as input, the encoded visual feature $\textbf{V}$ is obtained as:
\begin{equation} 
    \textbf{V} = f_v(V) \in \mathbb{R}^{T \times H \times W \times D_f} ,
\end{equation}
where $H$ and $W$ represent the spatial resolution of the downsampled feature map, and $D_f$ is the feature dimension.

For audio feature extraction, we design $f_a$ using a rectangular 2D CNN kernel, ensuring that each audio segment aligns with its corresponding visual frame. The kernel size is defined as:
\begin{equation} 
    K_w = \Bigl\lfloor \frac{W_a}{T} \Bigr\rfloor, \quad K_h = H_a ,
\end{equation}
where $K_w$ adjusts the receptive field to match the $T$-frame partitioning of the spectrogram. The audio feature $\textbf{A}$ is then obtained as:
\begin{equation} 
    \textbf{A} = f_a(A) \in \mathbb{R}^{T \times D_f} .
\end{equation}

% ------
% ------

\vspace{-3mm}
\paragraph{Positional Encoding} 

Since self-attention is permutation-invariant, incorporating positional encoding is essential to preserve temporal and spatial structures. However, $\textbf{A}$ contains only temporal information, whereas $\textbf{V}$ retains both spatial and temporal dimensions. 

To ensure consistent encoding across modalities, we define spatial positional encoding ${\rm Pos_s} \in \mathbb{R}^{T \times H \times W \times D_s}$ and temporal positional encoding ${\rm Pos_t} \in \mathbb{R}^{T \times D_t}$. Here, $D_s$ and $D_t$ denote the dimensions of spatial and temporal positional encodings, respectively, and $D_s = D_f$ ensures compatibility with visual features $\textbf{V}$. The final modality-specific feature representations are computed as:
\begin{align}
    \tilde{\textbf{V}} &= [\textbf{V} + {\rm Pos}_s ; {\rm Pos}_t] \in \mathbb{R}^{T \times H \times W \times D}, \\
    \tilde{\textbf{A}} &= [\textbf{A} ; {\rm Pos_t}] \in \mathbb{R}^{T \times D} .
\end{align}
where spatial positional encoding is added element-wise to visual features, while temporal encoding is concatenated across both modalities. As a result, the final dimensions of $\tilde{\textbf{V}}$ and $\tilde{\textbf{A}}$ are $D = D_f + D_t$, ensuring that both modalities are encoded in a spatiotemporally consistent representation.

\subsection{Audio-Spatial-Temporal Attention Block}
\label{approach:audio-spatial-temporal_attention_block}

To construct a 3D feature representation with spatial and temporal axes for the AST block's input, we first flatten the spatial dimensions $H \times W$ of $\tilde{\textbf{V}}$, ensuring that each timestamp $t$ retains $H \cdot W$ spatial positions. The processed visual feature is then concatenated with the audio feature:
\begin{equation}
    \textbf{Z}^0 = [\tilde{\textbf{A}}; \tilde{\textbf{V}}] \in {\mathbb R}^{T \times (1 + H \cdot W) \times D}.
    \label{eqn:3d_input}
\end{equation}
The AST Block consists of two key components: Spatial and Temporal Attention. Spatial Attention learns fine-grained cross-modal interactions between audio and visual features at each timestamp, ensuring a precise spatial alignment between the two modalities. Following this, Temporal Attention captures long-range dependencies along the time axis, enhancing the model’s ability to track temporal dynamics and maintain consistency across frames. The operations of the AST Block at layer $l$ are formulated as follows:
\begin{align}
    &\textbf{Y}^{l-1} = {\rm SpatialAttention}(\textbf{Z}^{l-1}), \\ 
    &\textbf{Z}^l = {\rm TemporalAttention}(\textbf{Y}^{l-1}).
\end{align}

Both Spatial and Temporal Attention are implemented using Multi-Head Self-Attention (MSA), with sequence lengths defined differently for each type. Spatial Attention applies self-attention across the $1 + H \cdot W$ dimension, capturing interactions between audio and visual features within a single frame. In contrast, Temporal Attention operates along the $T$ dimension, modeling dependencies over time. This structure allows each timestamp to encode both spatial and temporal relationships efficiently. As a result, these attentions can be implemented by simply adding a transpose operation at the end of the standard MSA computation. The detailed formulation of Spatial Attention is as follows:
\begin{align}
    \textbf{Y}_{s}^{l-1} &= {\rm MSA}_{\rm spatial}({\rm LN}(\textbf{Z}^{l-1})) + \textbf{Z}^{l-1}, \\
    \textbf{Y}_{s'}^{l-1} &= {\rm FFN}_{\rm spatial}({\rm LN}(\textbf{Y}_{s}^{l-1})) + \textbf{Y}_{s}^{l-1},\\
    \textbf{Y}^{l-1}  &= {\rm Transpose}(\textbf{Y}_{s'}^{l-1}, {\rm dim}=(1, 0)).
\end{align}

\subsection{Training Objective}
\label{approach:training_objective}

Our approach builds upon the cross-modal multiple-instance contrastive learning loss introduced in EZ-VSL~\cite{mo2022localizing}, but incorporates two key modifications. First, we introduce a temporal component to explicitly model time-dependent relationships. Second, we refine the optimization goal for negative bags to enhance training stability.

Given an AST Block with depth $L$, the final output $\textbf{Z}^L \in \mathbb{R}^{T \times (1+ H \cdot W) \times D}$ is decomposed into an audio representation $\hat{\textbf{A}} \in \mathbb{R}^{T \times D}$ and a visual representation $\hat{\textbf{V}} \in \mathbb{R}^{T \times H \times W \times D}$. The flattened visual features are restored to their original spatial structure $H \times W$.

Unlike EZ-VSL, which constructs a bag of visual features corresponding to a global audio representation, our approach incorporates temporal information by redefining visual bags at the frame level. Specifically, for a given timestamp $t$ in video $X$, the audio segmentation representation $\hat{\textbf{A}}^t$ is associated with a bag of visual features comprising all spatial locations within the corresponding visual representation $\hat{\textbf{V}}^t = \{\hat{\textbf{v}}^{t}_{xy} : \forall x \in [1,H], y \in [1,W]\}$.

To enforce alignment, $\hat{\textbf{A}}^t$ should exhibit high similarity with at least one instance in its positive bag, while maintaining low similarity across all instances in negative bags. We define the negative bags of $\hat{\textbf{A}}_i^t$ from video $X_i$ as the bag of visual features from the same timestamp $t$ in other videos $X_j$ within the batch $B$, where $j \neq i$.

The positive and negative responses for the final optimization objective are defined as follows:
\begin{align}
    {\rm p}^t_i &= \max_{\hat{\textbf v} \in \hat{\textbf V}^t_i} \langle \hat{\textbf A}_i^t, \hat{\textbf v} \rangle ,\\ 
    {\rm n}^t_{ij} &= \frac{1}{HW} \sum_{\hat{\textbf v} \in \hat{\textbf V}^t_j} \langle \hat{\textbf A}_i^t, \hat{\textbf v} \rangle, \quad \forall j \neq i.
\end{align}
Here, $\langle \cdot, \cdot \rangle$ denotes the cosine similarity operation, which measures the similarity between two vectors.

While EZ-VSL maximizes similarity for both positive and negative bags, we instead compute the mean similarity for negative bags. This prevents the loss function from being dominated by noisy or outlier instances, leading to a more stable optimization process that better reflects the overall distribution of negative samples.
The audio-to-visual alignment loss ${\cal L}_{a \rightarrow v}$ is formulated as:
\begin{equation}
    {\cal L}_{a \rightarrow v} = - {\mathbb E}_{t,i} \left[ \log \frac{\exp({\rm p}_i^t)}{\exp({\rm p}^t_i) + \sum_{j \neq i}^B \exp({\rm n}^t_{ij}) }\right].
\end{equation}

The final training objective combines both audio-to-visual (${\cal L}_{a \rightarrow v}$) and visual-to-audio (${\cal L}_{v \rightarrow a}$) alignments, where ${\cal L}_{v \rightarrow a}$ is symmetrically defined using negative response ${\rm n}_{ji}^t$:
\begin{equation}
    {\cal L} = {\cal L}_{a \rightarrow v} + {\cal L}_{v \rightarrow a}.
\end{equation}

During inference, direct vector similarity is used to generate an audio-visual localization map, computed as:
\begin{equation}
    {\rm s}^{t}_{i,xy} = \langle \hat{\textbf A}^t_i, \hat{\mathbf v}^{t}_{i,xy} \rangle.
\end{equation}

%% file: sec/5_experiments.tex
\section{Experiments}
\label{sec:experiments}

% ------
\input{table/scenarios_experiments}
% ------

% ------
\input{table/cross_event_experiments_ciou}
% ------

\subsection{Experimental Setup}
\paragraph{Cross-event}
To assess the model's adaptability to dynamic real-world situations, we introduce the Cross-event subset, where the sound source changes over time. This setting evaluates whether a model can accurately track shifting audio sources. Cross-event videos are selected from AVATAR by identifying instances where a new audio-visual category emerges that was absent in earlier frames. Examples are shown in \cref{fig:senarios}.

\vspace{-0.3cm}
\paragraph{Training Data}
The training dataset is based on VGGSound~\cite{chen2020vggsound}. To prevent data leakage, we identified and removed 1,249 overlapping videos between AVATAR and VGGSound’s training set. Additionally, as factors like fps can influence model training and inference settings, we selected videos with 200–300 frames to align with AVATAR’s criteria. From a pool of 40k videos, a subset of 10k was randomly sampled for training.

\vspace{-0.3cm}
\paragraph{Baselines}
We evaluate three baseline models trained on VGGSound: EZ-VSL~\cite{mo2022localizing}, SLAVC~\cite{mo2022SLAVC}, and SSL-TIE~\cite{liu2022exploiting}. These models perform inference on individual audio-image pairs, and we follow the same setting for evaluation. However, as they do not inherently model temporal dynamics, direct comparison with our method may not be entirely fair. To address this, we conduct additional experiments using varying audio window lengths (10s, 5s, 2s, 1s). Results indicate that shorter audio windows lead to a sharp performance drop or remain stable but degrade significantly in Cross-event.

\vspace{-0.3cm}
\paragraph{Evaluation Metrics}
For quantitative evaluation, we adopt Consensus Intersection over Union (CIoU) and Area Under Curve (AUC) as the primary metrics, following previous studies~\cite{mo2022localizing, mo2022SLAVC, liu2022exploiting}. Additionally, for Off-screen scenario, we use the pixel-level True Negative (TN) percentage. Prior studies applied frame-level min-max normalization and used a heatmap threshold of 0.5 for evaluation. However, this normalization ensures that the maximum value within each frame is always 1, inevitably resulting in regions exceeding the threshold. This assumption implicitly presumes that a sound-emitting object is always present within the frame, contradicting the Off-screen scenario. To address this, we set the threshold for each model based on the top 10\% of pixel values across all labeled frames.

\subsection{Experimental Results}
\subsubsection{Analysis of Cross-event Videos}

To assess the model’s adaptability to dynamic real-world environments, we evaluate its performance in the Cross-event setting, where the sound-emitting object changes over time. This conditions tests whether AVL models can effectively localize temporally varying audio sources.

Results in \cref{tab:cross_event_experiments_ciou} and \cref{tab:cross_event_experiments_auc} demonstrate a notable performance drop in baseline models on the Cross-event subset compared to the overall benchmark (Total). Specifically, existing methods exhibit declines of up to $-5.36$ CIoU and $-5.00$ AUC, indicating their inability to adapt to shifting audio sources. This limitation stems from their reliance on static mappings between global audio and single-frame visual representations, preventing effective tracking of temporally evolving sound sources. As a result, these models struggle to localize sound sources accurately when they change over time.

In contrast, by explicitly incorporating temporal information, our model enables more effective tracking of dynamic sound sources. The results confirm this advantage, as our model exhibits only a minimal performance drop ($\Delta=-0.33$ for CIoU, $\Delta=-0.37$ for AUC) in the Cross-event setting. This demonstrates that, rather than relying on static mappings, our model leverages temporal context, maintaining stable localization performance even in dynamic environments.

% ------
\begin{figure*}[!t]
    \centering
    \includegraphics[width=\textwidth]{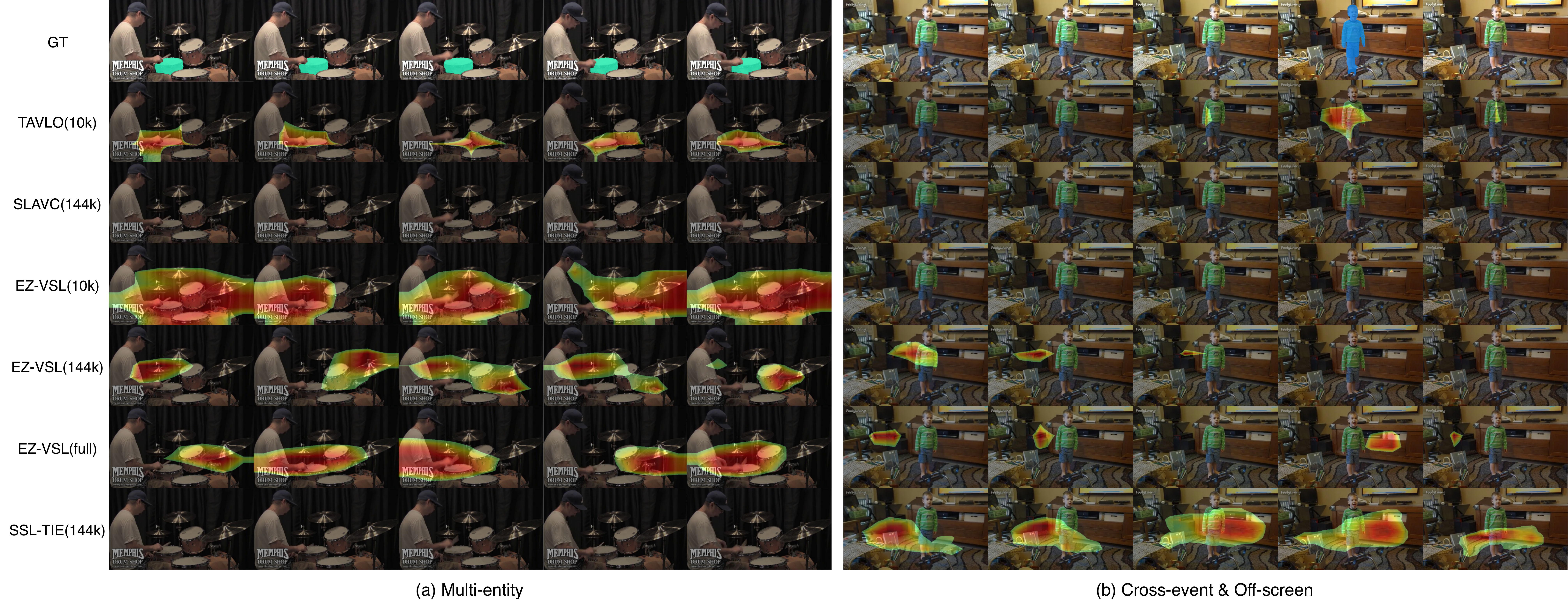}

    \vspace{-0.3cm}
    \caption{Qualitative Comparison. (a) Multi-entity: A drum performance scene where multiple drums are visible, but only the snare drum produces sound. (b) Cross-event \& Off-screen: A scenario where a woman speaks off-screen before a child begins speaking on-screen at a specific moment (4th frame).}
    \label{exp_vis:main_result}
    \vspace{-0.5cm}
\end{figure*}
% ------

% ------
\input{table/cross_event_experiments_auc}
% ------

\subsubsection{Scenario-wise Performance Evaluation}

To provide a more comprehensive assessment beyond conventional evaluation metrics, we conduct a scenario-based analysis to assess localization accuracy under diverse real-world conditions. Performance is evaluated at the frame level, considering only instances relevant to each scenario, as summarized in \cref{tab:scenario_experiments}.

In the Single-sound scenario, which represents the fundamental AVL task with a single sound-emitting instance, most models achieve high performance. Notably, SSL-TIE (144k) marginally outperforms our model by 0.15\% in AUC. However, apart from this case, our model surpasses all other baselines, achieving 13.42\% CIoU and 14.08\% AUC, demonstrating superior localization precision in single-source environments.

The Mixed-sound and Multi-entity scenarios introduce greater auditory and visual complexity, leading to significant performance degradation in baseline models. In contrast, our model achieves 14.13\% CIoU and 14.52\% AUC in the Mixed-Sound scenario, and 12.08\% and 12.69\% in the Multi-entity scenario, outperforming all baselines. Notably, despite not being explicitly designed for these conditions, our model maintains a performance similar to that in the Single-sound setting. This suggests that high-resolution temporal audio segment modeling, combined with a temporal-aware architecture leveraging motion cues, enhances its robustness. These results indicate that our approach effectively adapts to complex auditory environments and accurately differentiates multiple sound-emitting instances.

In the Off-screen scenario, TAVLO exhibits a relatively low TN(\%) value. However, this can be attributed to the fact that other models frequently produce false positives in non-Off-screen frames due to the top 10\% pixel thresholding of the heatmap. This suggests that their reduced false positive rate in Off-screen cases may stem from lower overall performance in other scenarios. To examine this effect, we re-evaluated performance using only videos with Off-screen frames, adjusting the heatmap threshold to the top 5\% of pixel values. The revised results $^\dagger$TN(\%) indicate that most models perform comparably under this condition.

\subsection{Qualitative Results}

\cref{exp_vis:main_result} compares audio-visual localization results from the proposed model and baseline methods in two challenging videos: (a) Multi-entity and (b) Cross-event \& Off-screen.

In \cref{exp_vis:main_result}(a), a drum performance scene features multiple drums, but only the snare drum produces sound. The key challenge is distinguishing the actual sound source from visually similar but silent objects. The proposed model consistently localizes the snare drum across all five frames, whereas baselines often highlight non-sounding drums, suggesting a reliance on visual similarity rather than true audio-visual correspondence. This demonstrates the proposed model’s ability to learn precise cross-modal associations.

\cref{exp_vis:main_result}(b) depicts a scene where an off-screen woman is speaking until a child within the frame begins speaking in the fourth frame. The challenge is to avoid false positives when the off-screen speaker is active and correctly localize the on-screen speaker at the right moment. The proposed model successfully adapts to this transition, accurately identifying the child while avoiding erroneous detections during off-screen speech. In contrast, baseline models, particularly EZ-VSL(144k), EZ-VSL(full), and SSL-TIE(144k), struggle due to their reliance on global audio features, which incorporate the child's speaking voice. As a result, these models often make incorrect predictions that do not align with the actual speaking moment. This highlights a fundamental limitation of existing models in capturing fine-grained temporal changes in sound localization.

%% file: table/scenarios_experiments.tex
\begin{table*}[t]
\begin{center}
\caption{Performance Comparison Across Scenarios. Quantitative results of various methods across different audio-visual localization scenarios: Single-sound, Mixed-sound, Multi-entity, and Off-screen.}
\vspace{-3mm}
\resizebox{\textwidth}{!}{
\renewcommand{\arraystretch}{1.25}
\footnotesize
\begin{tabular}{L{2.7cm}C{1.35cm}C{1.35cm}C{0.05cm}C{1.35cm}C{1.35cm}C{0.05cm}C{1.35cm}C{1.35cm}C{0.05cm}C{1.35cm}C{1.35cm}}
\Xhline{2.5\arrayrulewidth}
 & \multicolumn{2}{c}{(1) Single-sound} && \multicolumn{2}{c}{(2) Mixed-sound} && \multicolumn{2}{c}{(3) Multi-entity} && \multicolumn{2}{c}{(4) Off-screen}\\
\cmidrule(r){2-12}
\textbf{Method} & CIoU(\%)$\uparrow$ & AUC(\%)$\uparrow$ && CIoU(\%)$\uparrow$ & AUC(\%)$\uparrow$ && CIoU(\%)$\uparrow$ & AUC(\%)$\uparrow$ && TN(\%)$\uparrow$ & $^\dagger$TN(\%)$\uparrow$\\
\hline \hline
SLAVC(144k)~\cite{mo2022SLAVC} & \phantom{0}9.07 & 10.60 && \phantom{0}6.31 & \phantom{0}7.88 && \phantom{0}6.41 & \phantom{0}7.96 && 96.46 & 95.75 \\
EZ-VSL(10k)~\cite{mo2022localizing} & \phantom{0}9.66 & 11.07 && \phantom{0}8.16 & \phantom{0}9.35 && \phantom{0}6.87 & \phantom{0}8.32 && \textbf{96.91} & 95.58 \\
EZ-VSL(144k)~\cite{mo2022localizing} & 10.92 & 12.22 && \phantom{0}6.97 & \phantom{0}8.34 && \phantom{0}5.80 & \phantom{0}7.42 && 96.47 & \textbf{96.45}\\
EZ-VSL(full)~\cite{mo2022localizing} & 12.17 & 13.38 && \phantom{0}7.67 & \phantom{0}8.91 && \phantom{0}6.96 & \phantom{0}8.40 && 95.43 & 95.84 \\
SSL-TIE(144k)~\cite{liu2022exploiting} & 13.10 & \textbf{14.23} && \phantom{0}5.19 & \phantom{0}6.76 && \phantom{0}5.50 & \phantom{0}7.12 && 90.82 & 94.55 \\
\rowcolor{gray!10}
TAVLO(10k) & \textbf{13.42} & 14.08 && \textbf{14.13} & \textbf{14.52} && \textbf{12.08} & \textbf{12.69} && 91.18 & 95.02 \\
\Xhline{2.5\arrayrulewidth}
\multicolumn{12}{l}{\footnotesize $^\dagger$TN(\%) evaluated with a heatmap threshold, defined as the top 5\% highest pixel values, on videos that include off-screen scenarios.}
\end{tabular}
}

\label{tab:scenario_experiments}
\end{center}
\vspace{-0.8cm}
\end{table*}

%% file: table/cross_event_experiments_ciou.tex
\begin{table}[t]
\begin{center}
\caption{CIoU Performance on Cross-event Scenarios. Comparison of CIoU(\%) on the full AVATAR (Total) and Cross-event, with $\Delta$ indicating the performance drop.}
\vspace{-3mm}
\footnotesize
\resizebox{\linewidth}{!}{
\renewcommand{\arraystretch}{1.25}
\begin{tabular}{L{2.8cm}C{1.4cm}C{1.4cm}C{1.1cm}}
\Xhline{2.5\arrayrulewidth}
& Total &\multicolumn{2}{c}{Cross-event}\\
\cmidrule(r){2-4}
\textbf{Method} & CIoU(\%)$\uparrow$ & CIoU(\%)$\uparrow$ & $\Delta$\\
\hline \hline
SLAVC(144k)~\cite{mo2022SLAVC} & \phantom{0}8.12 & \phantom{0}4.78 & -3.34 \\
EZ-VSL(10k)~\cite{mo2022localizing} & \phantom{0}8.95 & \phantom{0}5.08 & -3.87\\
EZ-VSL(144k)~\cite{mo2022localizing} & \phantom{0}9.38 & \phantom{0}5.19 & -4.19 \\
EZ-VSL(full)~\cite{mo2022localizing} & 10.50 & \phantom{0}5.26 & -5.24 \\
SSL-TIE(144k)~\cite{liu2022exploiting} & 10.39 & \phantom{0}5.03 & -5.36 \\
\rowcolor{gray!10}
TAVLO(10k) & \textbf{13.37} & \textbf{13.04} & \textbf{-0.33} \\
\Xhline{2.5\arrayrulewidth}
\end{tabular}
}

\label{tab:cross_event_experiments_ciou}
\end{center}
\vspace{-0.4cm}
\end{table}

%% file: table/cross_event_experiments_auc.tex
\begin{table}[t]
\begin{center}
\caption{AUC Performance on Cross-event Scenarios. Comparison of AUC(\%) on the full AVATAR (Total) and Cross-event, with $\Delta$ indicating the performance drop.}
\vspace{-3mm}
\footnotesize
\resizebox{\linewidth}{!}{
\renewcommand{\arraystretch}{1.25}
\begin{tabular}{L{2.8cm}C{1.4cm}C{1.4cm}C{1.1cm}}
\Xhline{2.5\arrayrulewidth}
& Total &\multicolumn{2}{c}{Cross Event}\\
\cmidrule(r){2-4}
\textbf{Method} & AUC(\%)$\uparrow$ & AUC(\%)$\uparrow$ & $\Delta$\\
\hline \hline
SLAVC(144k)~\cite{mo2022SLAVC} & \phantom{0}9.67 & \phantom{0}6.66 & -3.07 \\
EZ-VSL(10k)~\cite{mo2022localizing} & 10.33 & \phantom{0}6.73 & -3.60 \\
EZ-VSL(144k)~\cite{mo2022localizing} & 10.74 & \phantom{0}6.86 & -3.88 \\
EZ-VSL(full)~\cite{mo2022localizing} & 11.75 & \phantom{0}7.04 & -4.71 \\
SSL-TIE(144k)~\cite{liu2022exploiting} & 11.68 & \phantom{0}6.68 & -5.00 \\
\rowcolor{gray!10}
TAVLO(10k) & \textbf{13.98} & \textbf{13.61} & \textbf{-0.37} \\
\Xhline{2.5\arrayrulewidth}
\end{tabular}
}

\label{tab:cross_event_experiments_auc}
\end{center}
\vspace{-0.4cm}
\end{table}

%% file: sec/6_conclusion.tex
\section{Conclusion}
\label{sec:conclusion}

This study highlights the critical role of temporal modeling in audio-visual localization and introduces a new evaluation standard for Video-centric AVL. While prior work has largely focused on static frame-level associations, our findings demonstrate that capturing temporal dynamics is essential for accurate and robust sound source localization. To this end, we propose a novel benchmark and model, providing a systematic framework for evaluating AVL performance in dynamic environments and emphasizing the importance of precise multimodal alignment over time.

Experimental results reveal the limitations of static image-based approaches, underscoring the necessity of video-centric methodologies in audio-visual perception. Notably, integrating temporal context significantly enhances model performance, offering valuable insights for future research in multimodal learning, fine-grained object differentiation, and localization in complex acoustic scenes.

Despite its contributions, this study has limitations. It assumes at least partial alignment between audio and visual elements, leaving the independent localization of off-screen sounds as an open challenge. Additionally, while the benchmark introduces diverse evaluation scenarios, it does not prescribe specific methods for optimizing performance in each case. Future research should address Video-centric AVL strategies for off-screen sounds and develop scenario-specific learning frameworks.

%% file: sec/acknowledgements.tex
\section*{Acknowledgements}
This work was supported by NRF (2021R1A2C3006659) and IITP grants (RS-2021-II211343, RS-2022-II220320) , all funded by the Korean Government.